\documentclass[preprint,sort&compress,11pt,3p]{elsarticle}
\usepackage{moreverb}
\usepackage{mathtools}
\usepackage{amsmath}
\usepackage{amssymb}
\usepackage{algorithmic}
\usepackage{graphics}
\usepackage{graphicx}
\usepackage{subfigure}
\usepackage{caption}
\usepackage{enumitem}
\usepackage{extarrows}
\usepackage{color}
\usepackage{framed}
\usepackage{wrapfig}
\usepackage{bm}
\usepackage{mathrsfs}
\usepackage{mathabx}
\usepackage{multirow}
\usepackage{longtable}
\usepackage{hyperref}
\usepackage{paralist}
\usepackage{indentfirst}
\usepackage{relsize}
\usepackage{extarrows}
\usepackage{lineno,xcolor}
\usepackage{upgreek}
\usepackage{bm}
\usepackage{mwe}
\usepackage{xcolor}
\usepackage{booktabs}
\usepackage[flushleft]{threeparttable}
\usepackage[linesnumbered,lined,ruled]{algorithm2e}

\newcommand{\eref}[1]{(\ref{#1})}

\hypersetup{
bookmarks=true,
bookmarksopen=true,
bookmarksnumbered=true,
unicode=false,
pdftoolbar=true,
pdfmenubar=true,
pdffitwindow=false,
pdfstartview={FitH},
pdftitle={My title},
pdfauthor={Author},
pdfsubject={Subject},
pdfcreator={Creator},
pdfproducer={Producer},
pdfkeywords={keywords},
pdfnewwindow=true,
colorlinks=true,
linkcolor=blue,
citecolor=blue,
filecolor=magenta,
urlcolor=blue
}

\journal{Journal of Computational Physics}

\begin{document}
	
\title{\Large Physics-informed Deep Super-resolution for Spatiotemporal Data}

\author[NU1]{Pu Ren}
\ead{ren.pu@northeastern.edu}
\author[NU2]{Chengping Rao}
\author[NU2]{Yang Liu}
\author[SU]{Zihan Ma}
\author[NU1]{Qi Wang}
\author[ND]{Jian-Xun Wang}
\author[RUC1,RUC2]{Hao Sun\corref{cor}}
\ead{haosun@ruc.edu.cn}
\cortext[cor]{Corresponding author.}

\address[NU1]{Department of Civil and Environmental Engineering, Northeastern University, Boston, MA 02115, USA}
\address[NU2]{Department of Mechanical and Industrial Engineering, Northeastern University, Boston, MA 02115, USA}
\address[SU]{West China School of Public Health, Sichuan University, Chengdu, Sichuan 610041, China}
\address[ND]{Department of Aerospace and Mechanical Engineering, University of Notre Dame, Notre Dame, IN 46556, USA}
\address[RUC1]{Gaoling School of Artificial Intelligence, Renmin University of China, Beijing, 100872, China}
\address[RUC2]{Beijing Key Laboratory of Big Data Management and Analysis Methods, Beijing, 100872, China}

\begin{abstract}
	\small
High-fidelity simulation of complex physical systems is exorbitantly expensive and inaccessible across spatiotemporal scales. Recently, there has been an increasing interest in leveraging deep learning to augment scientific data based on the coarse-grained simulations, which is of cheap computational expense and retains satisfactory solution accuracy. However, the major existing work focuses on data-driven approaches which rely on rich training datasets and lack sufficient physical constraints. To this end, we propose a novel and efficient spatiotemporal super-resolution framework via physics-informed learning, inspired by the independence between temporal and spatial derivatives in partial differential equations (PDEs). The general principle is to leverage the temporal interpolation for flow estimation, and then introduce convolutional-recurrent neural networks for learning temporal refinement. Furthermore, we employ the stacked residual blocks with wide activation and sub-pixel layers with pixelshuffle for spatial reconstruction, where feature extraction is conducted in a low-resolution latent space. Moreover, we consider hard imposition of boundary conditions in the network to improve reconstruction accuracy. Results demonstrate the superior effectiveness and efficiency of the proposed method compared with baseline algorithms through extensive numerical experiments.   
\end{abstract}


\maketitle

\section{Introduction}\label{s:intro}
Modeling high-resolution (HR) response of nonlinear dynamical systems (i.e., at sufficiently small spatiotemporal scales) has emerged as one of the dominant research pursuits in the field of scientific computing, thanks to its capability of providing insights into physical quantities of interest and reliable assessments for operational planning. For instance, numerical simulation of turbulence at fine grids allows for unveiling the richness of its varieties and portraying the delicately interactive behaviors of turbulent flows. Although scientists have developed valid governing equations to describe these physical systems, simulating such high-fidelity complex PDEs requires massive computational efforts, meanwhile posing significant challenges for storage space. With the limitation of computational and experimental resources, it is rather typical to store the spatiotemporally-downscaled data for post hoc analysis. This pressing issue calls for effective data augmentation approaches in scientific modeling for recovering the HR solutions from low-resolution (LR), e.g., sparse and incomplete data.

In order to upsample the LR scientific data to HR full-field dynamics, we refer to the idea of super-resolution (SR) in image and video processing. The image and video SR tasks are prohibitively ill-posed inverse problems considering the unknown blur and noise, the insufficient constraints and the non-uniqueness of reconstructed solutions. Previously, a seminal of breakthrough on single-image SR (SISR) has been made based on sparse signal representation \cite{yang2010image}, which is mathematically elegant and robust. Furthermore, due to the rapid development of deep learning (DL), there is another milestone work SRCNN \cite{dong2015image} remarkably improving the performance of SISR by firstly introducing deep neural networks (DNNs). Later, large amounts of network architectures have been proposed and continuously achieved huge success in this intractable task, mainly covering the variants of generative adversarial networks (GAN) \cite{ledig2017photo,zhang2019ranksrgan,chan2021glean}, residual blocks \cite{kim2016accurate,lim2017enhanced,yu2018wide,zhang2018image} and attention scheme \cite{yang2020learning,chen2021pre}. In the context of video SR, apart from the spatial upsampling, learning optical flow is another core and challenging issue. Many investigations leverage linear frame interpolation \cite{niklaus2017video,liu2017video,jiang2018super}, quadratic video interpolation \cite{xu2019quadratic,liu2020enhanced} and phase-based interpolation \cite{meyer2018phasenet} for modeling temporal dependency and estimating multiple intermediate frames. Besides, the recurrent convolutional neural network is also proven to be an effective tool for video SR \cite{sajjadi2018frame,xiang2020zooming}.

Thanks to the great success in image \cite{yang2010image,dong2015image,shi2016real,lim2017enhanced,ledig2017photo,yu2018wide,zhang2018image,chen2021pre} and video super-resolution (SR) \cite{niklaus2017video,sun2018pwc,jiang2018super,niklaus2018context,wang2019edvr,xiang2020zooming}, a considerable studies have grown up around the interplay between physics and machine learning (ML), with respect to SR of scientific data. The majority of existing investigations mainly cluster in purely data-driven ML methods which approximate the HR solutions based on potentially noisy and under-resolved simulations in turbulence \cite{xie2018tempogan,bar2019learning,liu2020deep,fukami2021machine,kochkov2021machine}, smoke \cite{bai2020dynamic} and climate \cite{stengel2020adversarial} modeling. Nevertheless, different from general SR tasks for images and videos, scientific data augmentation has more explicit and well-defined objectives (e.g., obeying underlying governing equations), which could facilitate the convergence and generalization abilities with the physical principles incorporated into network architectures. There exist two primary branches of DL models informed by physics for dynamical systems: continuous learning \cite{raissi2019physics,raissi2020hidden,lu2021learning,rao2021physics,karniadakis2021physics,chen2021physics} and discrete learning \cite{long2018hybridnet,zhu2019physics,geneva2020modeling,wang2020towards,zhang2020physics,zhang2020physics2,gao2021phygeonet,rao2021embedding,ren2022phycrnet,gao2022physics,liu2022predicting}. The continuous learning scheme typically employs fully-connected neural networks (NNs), which is excellent in capturing global solution patterns but generally limited to low-dimensional PDE systems. The discrete learning leverages convolutional/graph NNs for modeling nonlinear dynamics, projected on mesh grids, which is capable of well portraying the local solution details. In addition, there are many other significant approaches for learning nonlinear dynamics, including ``white-box'' models by embedding physics into the network \cite{long2018pde,long2019pde,rao2021embedding,rao2022discovering} and neural operators \cite{li2020neural,li2020fourier,lu2021learning}.

Therefore, more recently, a flurry of work based on physics-informed DL has been recently developed for SR of scientific data \cite{han2019tsr,bode2019using,wang2020physics,subramaniam2020turbulence,esmaeilzadeh2020meshfreeflownet,gao2021super}. Currently, the majority of physics-informed learning approaches for SR are limited to spatial super-resolution (S-SR) tasks (e.g., steady-state problems \cite{wang2020physics,subramaniam2020turbulence,gao2021super}), or only consider temporal super-resolution (T-SR) \cite{han2019tsr}. A notable latest study explored the potential of 3D convolutional operations for spatiotemporal super-resolution (ST-SR) \cite{esmaeilzadeh2020meshfreeflownet}, which exhibits excellent performance for 2D Rayleigh-B{\'e}nard convection (RBC) system. In this paper, we present a novel and effective ST-SR method for general spatiotemporal systems via physics-informed discrete learning. Distinct from directly using 3D convolution, we explicitly preserve the temporal evolution structure by decomposing the entire ST-SR into two modules, i.e., T-SR and S-SR. This decomposition is inspired by the observation that the temporal derivatives and spatial terms are independent in a general form of PDEs as illustrated in Section \ref{s:method}. This separable scheme is conducive to computational efficiency, as well as promotes flexibility and scalability to high-dimensional PDE systems (e.g., 3D systems).

Overall, our methodological contributions include: (1) a novel network architecture of physics-informed deep SR (PhySR) for scientific data, and (2) an intrinsic collaboration between high-order finite difference-based spatiotemporal filtering and hard-encoding of boundary conditions (BCs). Specifically, we apply interpolation to estimate the temporally upsampled solution, and further leverage a convolutional-recurrent network (i.e., ConvLSTM \cite{shi2015convolutional}) for learning temporal refinement and dynamical evolution on low-dimensional latent features during the T-SR process. In the second-stage (i.e., S-SR), we employ the stacked residual blocks with wide activation \cite{yu2018wide} to extract the spatial features and sub-pixel layers with pixelshuffle \cite{shi2016real} for spatial upsampling. In addition, a global residual connection from the temporally-interpolated results to spatially-reconstructed variables is considered. Furthermore, the numerical differentiation determines the PDE derivative terms which are further utilized to construct PDE residual loss, and hard-imposition of BCs eliminates the risk of sacrificing solution accuracy on physical boundaries. Comprehensive numerical experiments on different types of PDE systems have been implemented for evaluating the performance of our proposed approach, compared with two baseline algorithms. Moreover, we validate the effectiveness of the significant modules of our PhySR by conducting an ablation study on ConvLSTM, the introduction of physics loss and the encoding of BCs.

The rest of the paper is organized as follows. Section \ref{s:prob} sets up the problem of ST-SR of spatiotemporal scientific data. Section  \ref{s:method} describes the network structure of two neural modules (i.e., T-SR and S-SR), hard-encoding strategy of BCs and physics-informed optimization. In Section \ref{s:experiment}, we show the extensive experiments between our proposed framework and baseline methods. In Section \ref{s:discussion}, we discuss the advantages, the limitations of our current work and the outlook of future directions. Section \ref{s:conclusion} concludes the entire paper.

\section{Problem setup}\label{s:prob}

\begin{figure}[t]
\centering
\includegraphics[width=0.99\columnwidth]{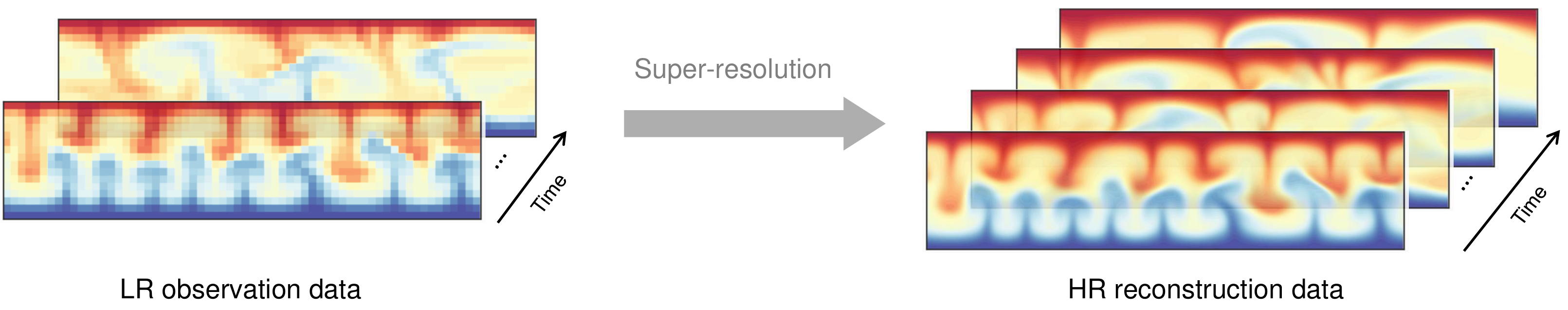}
\caption{Schematic of SR for 2D Rayleigh-B{\'e}nard convection system with respect to temperature. The LR observation data shown in the left part is measured at coarse-grid, and the HR reconstruction is based on fine-mesh with more time snapshots. Note that the spatial and temporal discretization considered here are both in uniform mode.}
\label{fig:prob_setup}
\end{figure}

Spatiotemporal dynamical systems are typically governed by PDEs. Let us consider a set of $m$-dimensional parametric PDEs in a general form:
\begin{equation}
    \label{eq:pde}
    \mathbf{u}_t = \mathcal{F}[\mathbf{x}, t, \mathbf{u}, \mathbf{u}_x, \mathbf{u}_y, \nabla \mathbf{u}, \cdots;\boldsymbol{\lambda}],
\end{equation}
where $\mathbf{x}$ and $t$ represent spatial and time coordinates, respectively. $\mathbf{u}(\mathbf{x},t)$ is the system solution comprised of $n$ state variables and the subscripts denote the partial derivatives, e.g., $\mathbf{u}_x=\partial \mathbf{u}/\partial x$. $\nabla$ denotes the Nabla operator (e.g., with the formulation of $[\partial/\partial x,\partial/\partial y]^\top$ when $m=2$). Moreover, $\mathcal{F}[\cdot]$ is the nonlinear functional consisting of polynomial and derivative terms and parameterized by $\boldsymbol{\lambda}$. The specific initial and boundary conditions (I/BCs) are defined as $\mathcal{I}(\mathbf{u},\mathbf{u}_t;t=0,\mathbf{x} \in \Omega)=0$ and $\mathcal{B}(\mathbf{u},\nabla\mathbf{u}, \cdots;\mathbf{x} \in \partial \Omega)=0$ respectively, where $\partial \Omega$ is the boundary of the physical domain. Given LR data $\mathbf{u}^{l}$, our objective is to spatiotemporally recover the HR variables $\widehat{\mathbf{u}}^{h}$ with consistent dynamical patterns, as shown in Figure \ref{fig:prob_setup}. Two constraints are considered in this paper to solve this ill-posed task: (1) reconstruction constraint, which requires the recovered HR data $\widehat{\mathbf{u}}^{h}$ corresponding to the LR input measurement data $\mathbf{u}^{l}$ with respect to the observed dynamical model; (2) physics prior, which assumes that LR and HR scientific data both obey the underlying governing equations.

\section{Methodology}\label{s:method}

\begin{figure}[t]
    \centering
    \includegraphics[width=0.99\textwidth]{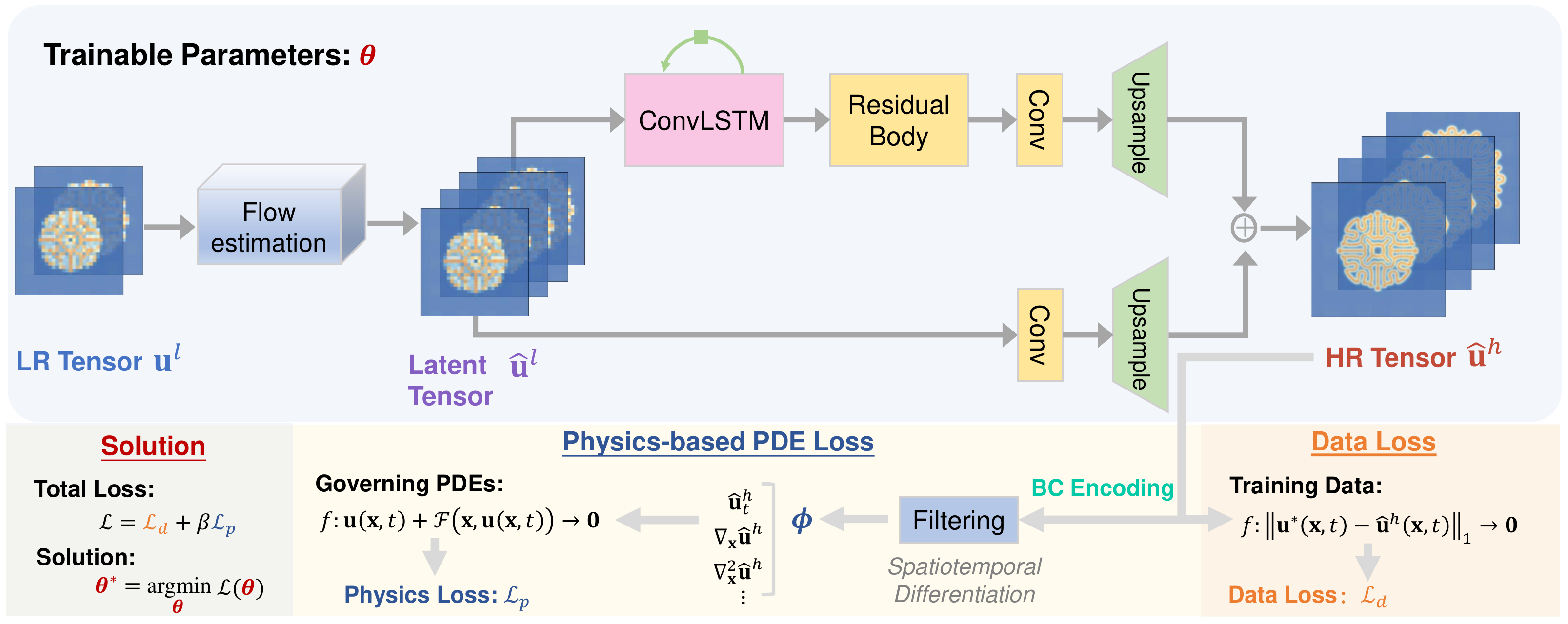} 
    \caption{Overview of the proposed PhySR framework. \textbf{Upper:} the network construction. We generate the intermediate frames by using temporal interpolation as flow estimation, and apply ConvLSTM for temporal refinement. To reconstruct the spatial resolution, we utilize the stacked residual blocks for feature extraction and sub-pixel layers with pixelshuffle for upsampling. Moreover, a global residual connection (i.e., in the lower arrow ``Conv+Upsample'') is considered here from the estimated LR tensor $\widehat{\mathbf{u}}^l$. \textbf{Bottom:} the loss components. Both the PDE residual loss $\mathcal{L}_p$ and the data loss $\mathcal{L}_d$ are computed based on the HR tensor $\widehat{\mathbf{u}}^h$. Here the BCs are hard encoded into the network when calculating PDE derivatives through finite difference filtering.}
    \label{fig:framework}
\end{figure}

An overview of our proposed framework of PhySR is presented in Figure \ref{fig:framework}. Based on the observation that time derivatives and spatial terms are mathematically separated, we design the network mainly consisting of the temporal upsampling (i.e., interpolation and refinement), as well as the spatial reconstruction. We first upscale the LR measurements to the desired frame rate in time dimension using the interpolation, and then apply ConvLSTM layers for refining the temporal approximation. Afterwards, for each snapshot, we leverage the residual blocks with wide activation and pixelshuffle layers for spatial upsampling. It is worthwhile to mention that all of the feature extraction processes are conducted in the LR latent space. The entire framework is ``end-to-end''. In addition, we define the upscaling factors of time snapshots and spatial resolution as $r_t$ and $r_s$ respectively. In this section, we take two LR input frames $\{\mathbf{u}_{k-1}^l,\mathbf{u}_{k+1}^l\}$ at time $k-1$ and $k+1$ for illustration (i.e., $r_t=2$), and attempt to recover the corresponding HR variables $\{\widehat{\mathbf{u}}_{k-1}^h,\widehat{\mathbf{u}}_{k+1}^h\}$ as well as synthesize the intermediate HR frame $\widehat{\mathbf{u}}_{k}^h$ simultaneously.

\subsection{ConvLSTM}\label{s:convlstm}
Firstly, we introduce the critical neural component (i.e., ConvLSTM) in our PhySR framework, especially for temporal upsampling. ConvLSTM is an extension from long short-term memory (LSTM) architecture \cite{hochreiter1997long,graves2013generating,sutskever2014sequence} and specifically designed for learning spatiotemporal sequence-to-sequence models. It is noteworthy that LSTM is a variant of recurrent neural network (RNN) with the purpose of mitigating the issues of gradient vanishing and explosion. Moreover, a major advantage of ConvLSTM is that it works as an implicit and nonlinear numerical scheme for modeling dynamical systems \cite{ren2022phycrnet}.

\begin{figure}[t]
\centering
\includegraphics[width=0.65\columnwidth]{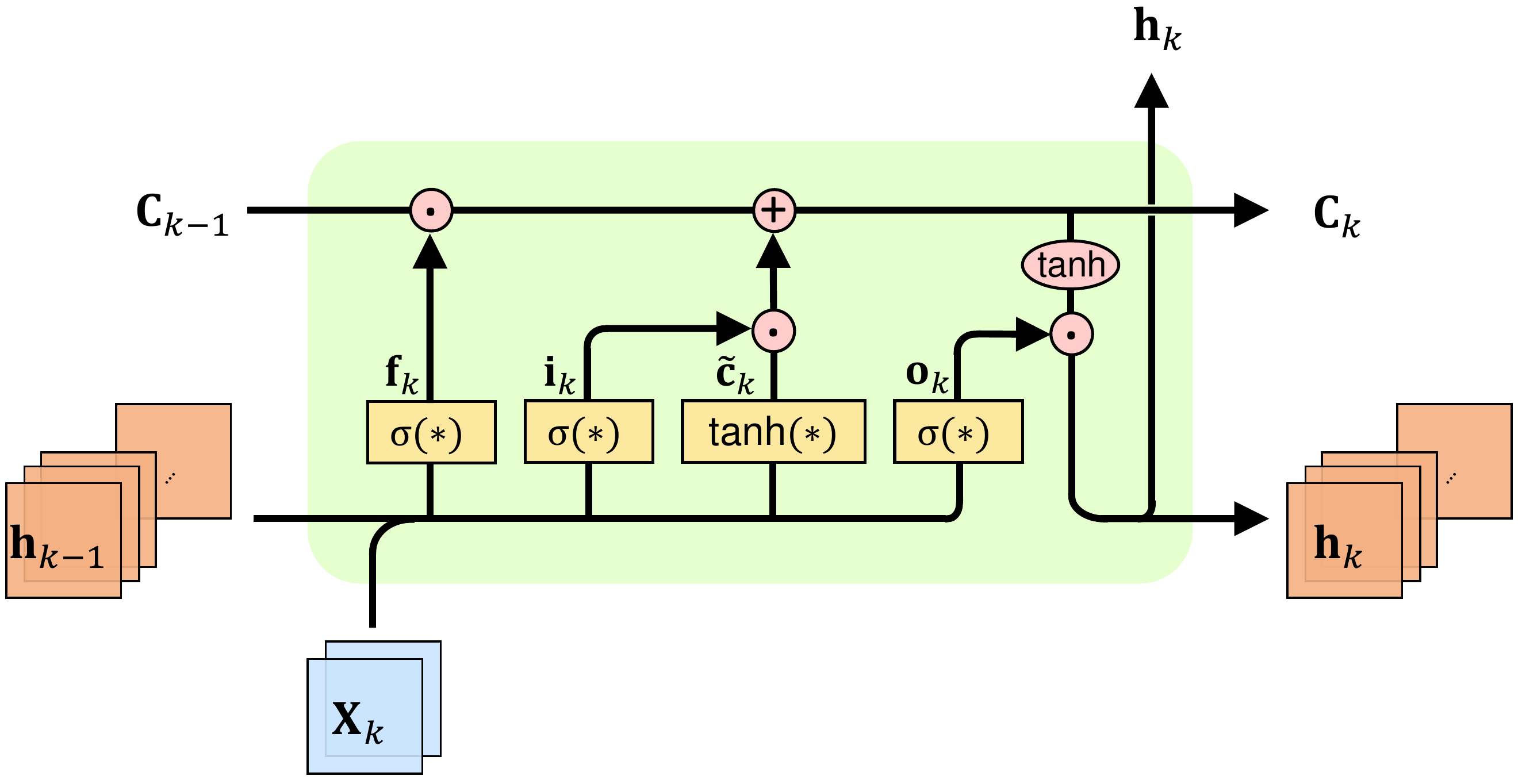}
\caption{Schematic of the ConvLSTM cell at time $t_k$. $\mathbf{X}_k$ is the input variable. $\{\mathbf{h}_k,\mathbf{c}_k\}$ are hidden state and cell state, respectively.}
\label{fig:convlstm}
\end{figure}

The architecture of ConvLSTM cell at time $t_k$ is presented in Figure \ref{fig:convlstm}, including the gate scheme and memory cells. The general principle is to update the hidden state $\mathbf{h}_k$ and cell state $\mathbf{C}_k$ at time $t_k$ by learning the transformation from the previous state variables $\{\mathbf{h}_{k-1},\mathbf{C}_{k-1}\}$ and the current input variable $\mathbf{X}_k$. Four gates control the information flow for input-to-state and state-to-state transitions, including a forget gate $\mathbf{f}_k$, an input gate $\mathbf{i}_k$, an internal cell $\widetilde{\mathbf{C}}_k$ and an output gate $\mathbf{o}_k$. In specific, the input and forget gates $\{\mathbf{i}_k,\mathbf{f}_k\}$ adaptively keep and clear the memory in $\mathbf{C}_{k-1}$. Moreover, a new cell candidate $\widetilde{\mathbf{C}}_k$ is generated by synthesizing the information from input variable $\mathbf{X}_k$ and previous hidden state $\mathbf{h}_{k-1}$. They all contribute to forming the updated cell state $\mathbf{C}_k$ at the current time step $t_k$. Lastly, the hidden state $\mathbf{h}_k$ is further updated by regulating the cell state $\mathbf{C}_k$ based on the output gate $\mathbf{o}_k$. The analytical formulations for four gate operations and state variables are given by: 
\begin{equation}
    \begin{split}
        \label{eq:convlstm}
        \mathbf{i}_k &= \sigma (\mathbf{W}_{i} * [\mathbf{X}_k,\mathbf{h}_{k-1}] + \mathbf{b}_i), \quad \quad
        \mathbf{f}_k = \sigma (\mathbf{W}_{f} * [\mathbf{X}_k,\mathbf{h}_{k-1}] + \mathbf{b}_f), \\ 
        \widetilde{\mathbf{C}}_{k-1} &= \text{tanh}(\mathbf{W}_{c} * [\mathbf{X}_k,\mathbf{h}_{k-1}] + \mathbf{b}_c), \quad  
        \mathbf{C}_k = \mathbf{f}_k \odot \mathbf{C}_{k-1} + \mathbf{i}_k \odot \widetilde{\mathbf{C}}_{k-1}, \\
        \mathbf{o}_k &= \sigma (\mathbf{W}_{o} * [\mathbf{X}_k,\mathbf{h}_{k-1}] + \mathbf{b}_o), \quad \quad
        \mathbf{h}_k = \mathbf{o}_k \odot \text{tanh}(\mathbf{C}_k), \\ 
    \end{split}
\end{equation}
where tanh$(*)$ and $\sigma(*)$ represent hyperbolic tangent activation and sigmoid functions, respectively. Here $\odot$ and $*$ denote Hadamard product and the convolution operation. In addition, the weight and bias parameters of four gates in the ConvLSTM cell are correspondingly defined as $\{\mathbf{W}_i,\mathbf{W}_f,\mathbf{W}_c,\mathbf{W}_o\}$ and $\{\mathbf{b}_i,\mathbf{b}_f,\mathbf{b}_c,\mathbf{b}_o\}$.

\subsection{Pixel shuffle}\label{s:pixelshuffle}
\begin{figure}[t]
\centering
\includegraphics[width=0.6\columnwidth]{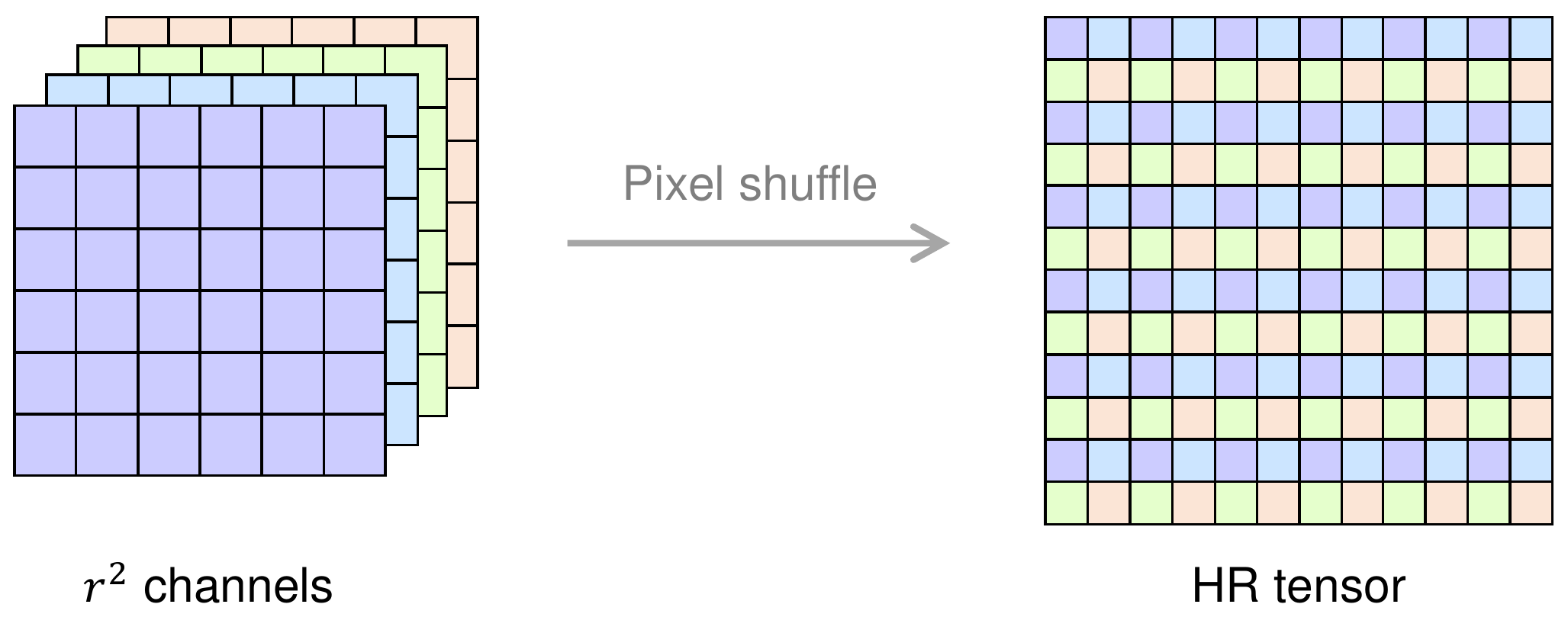}
\caption{Schematic of a pixel shuffle layer in 2D setting.}
\label{fig:pixelshuffle}
\end{figure}

Secondly, we leverage pixel shuffle to upscale LR latent features to HR variables, which is an efficient operation to conduct the sub-pixel convolutions \cite{shi2016real}. The graphic illustration is shown in Figure \ref{fig:pixelshuffle}. Let us consider an example of the pixel shuffle operation in 2D setting. We define a LR feature map with the tensor shape of $(C\times r^2,H,W)$ where $\{r,C,H,W\}$ denote upscaling factor, channel size, height and width respectively. The pixel shuffle reforms the elements in the LR feature map to a HR tensor with a larger spatial resolution of $(H\times r,W\times r)$ and a shrunken feature size of $C$. Although it is a simple operation, the pixel shuffle still exhibits excellent SR performance in image and video tasks \cite{shi2016real}. The second advantage is that it does not require huge computational efforts because the NN training before the spatial upsampling layer is based on the LR latent space. Furthermore, compared with other upscaling approaches (e.g., deconvolution), it leads to fewer checkerboard artifacts \cite{odena2016deconvolution}. Overall, pixel shuffle is selected as the spatial upsampling strategy in this paper considering its contribution to computational and memory efficiency. 

\subsection{Temporal upsampling}
In the temporal upsampling part, given the LR frames $\{\mathbf{u}_{k-1}^l,\mathbf{u}_{k+1}^l\}$, we first estimate the flow evolution to generate the intermediate LR frame $\widehat{\mathbf{u}}_k^l$ and then map them into the LR latent space as $\{\mathbf{h}_{k-1},\mathbf{h}_{k},\mathbf{h}_{k+1}\}$. Herein, we hybrid the traditional interpolation approaches and the deep convolutional-recurrent neural networks for initial flow estimation and temporal correction, respectively. The reason behind using interpolation is that we find it can effectively reduce the artifacts and oscillations in the time dimension, which is similarly reported in image SR \cite{odena2016deconvolution}. 

\subsubsection{Temporal interpolation}
Different from the general video frame estimation, scientific data SR is equation-specific which promisingly inspires the direction of temporal interpolation model. For instance, for those PDEs with the first-order time derivative (i.e., $\mathbf{u}_t$), applying linear interpolation is capable of roughly approximating the slow-evolutionary dynamics. Besides, previous methods \cite{niklaus2017video,xiang2020zooming} also implement the interpolation on the learned features to synthesize the intermediate feature maps considering the complex motions in the real world. Nevertheless, here we can directly interpolate on the input LR frames considering the explicit and simple time derivative terms, which remarkably improves the the computational efficiency. Furthermore, the following ConvLSTM layers can capture more straightforward and essential information from the input spatiotemporal dynamics compared with the learned features. Additionally, it helps to build a reliable global residual connection from the interpolated results and the final super-resolved variables. 

Let us define a generic form of the interpolation function $f[\cdot]$, which only works for temporal upscaling. Therefore, the initial flow estimation $\widehat{\mathbf{u}}_k^l$ is formulated as:
\begin{equation}
    \label{eq:interpolate}
    \widehat{\mathbf{u}}_k^l = f[\mathbf{u}_{k-1}^l,\mathbf{u}_{k+1}^l].
\end{equation}

\subsubsection{Temporal refinement}
Based on the direct estimation from the temporal interpolation, we further improve the capability of synthesizing intermediate frames by introducing DNNs for learning correction. Here we consider the ConvLSTM architecture, which is a spatiotemporal sequence-to-sequence learning tool for modeling long-term temporal dependencies. Given the consecutive LR interpolated frames $\{\mathbf{u}_{k-1}^l,\widehat{\mathbf{u}}_k^l,\mathbf{u}_{k+1}^l\}$, we aim to apply ConvLSTM to obtain the corresponding LR latent features $\{\mathbf{h}_{k-1},\mathbf{h}_k,\mathbf{h}_{k+1}\}$. Hence, the temporal updating at time $t_k$ is given by
\begin{equation}
    \label{eq:correction}
    \mathbf{h}_k, \mathbf{c}_k = \mathcal{NN}^{t}[\mathbf{h}_{k-1},\mathbf{c}_{k-1},\widehat{\mathbf{u}}_{k}^l;\boldsymbol{\theta}^t],
\end{equation}
where $\mathbf{h}_k$ and $\mathbf{c}_k$ are hidden state and cell state at time $t_k$; $\mathcal{NN}^{t}$ represents ConvLSTM networks and $\boldsymbol{\theta}^t$ is the corresponding learnable parameters. Herein, the temporal refinement is implicitly captured through two previous states (i.e., $\mathbf{h}_{k-1}$ and $\mathbf{c}_{k-1}$) and the current interpolated frame $\widehat{\mathbf{u}}_{k}^l$. Essentially, the learning process of Eq.\eref{eq:correction} can be regarded as a forward-propagated state-space model with nonlinear activation function. Therefore, due to the similarity to the numerical scheme, ConvLSTM can effectively extract the dynamics from input LR  measurements. Additionally, setting ConvLSTM on LR space can relieve the memory burden and facilitate the training efficiency.

\subsection{Spatial reconstruction}
After acquiring the sequence of LR feature maps $\{\mathbf{h}_{k-1},\mathbf{h}_k,\mathbf{h}_{k+1}\}$, we propose to reconstruct the corresponding HR snapshots $\{\widehat{\mathbf{u}}_{k-1}^h,\widehat{\mathbf{u}}_k^h,\widehat{\mathbf{u}}_{k+1}^h\}$ by applying a temporally-shared fully-convolutional network. Generally, there are two parts for spatial upsampling. The first neural component is the stacked residual blocks with wide activation to extract the deep spatial features and one sub-pixel layer with pixelshuffle for upsampling, which contributes to increasing the spatial resolution fast and efficiently. Here we define the HR output at time $t_k$ as $\mathcal{NN}^{s}[\mathbf{h}_k;\boldsymbol{\theta}^s]$.

Moreover, we consider adding a global residual connection directly from each interpolated LR snapshot $\widehat{\mathbf{u}}_k^l$ to the corresponding HR output $\mathcal{NN}^{r}[\widehat{\mathbf{u}}^l_k;\boldsymbol{\theta}^{r}]$, where single convolution layer and one sub-pixel layer are utilized for increasing the number of channels to $nr_s^m$ and further generating the HR counterparts respectively. Herein, $\mathcal{NN}^{s}[\cdot]$ and $\mathcal{NN}^{r}[\cdot]$ represent the first network component and the global residual network separately. $\boldsymbol{\theta}^s$ and $\boldsymbol{\theta}^{r}$ are the corresponding trainable parameters. Therefore, the final synthesized HR output variable $\widehat{\mathbf{u}}_k^h$ is formulated as
\begin{equation}
    \label{eq:sr}
    \widehat{\mathbf{u}}_k^h = \mathcal{NN}^{s}[\mathbf{h}_k;\boldsymbol{\theta}^s] + \mathcal{NN}^{r}[\widehat{\mathbf{u}}^l_k;\boldsymbol{\theta}^r].
\end{equation}
In addition, weight normalization \cite{salimans2016weight} is employed in this work for alleviating the training difficulty.

\subsection{Hard-encoding of BCs}
Basically, SR is a typical ill-posed task due to the harsh requirement of extrapolation from limited measurement data. However, prior knowledge is always accessible for SR of scientific data, including the specific governing equations and the I/BCs. The physical information can provide accurate description of the underlying dynamics and effective guidance for extrapolation. Therefore, here we propose to hard impose the BCs into the network to facilitate network training. 

\begin{figure}[t]
\centering
\includegraphics[width=0.5\columnwidth]{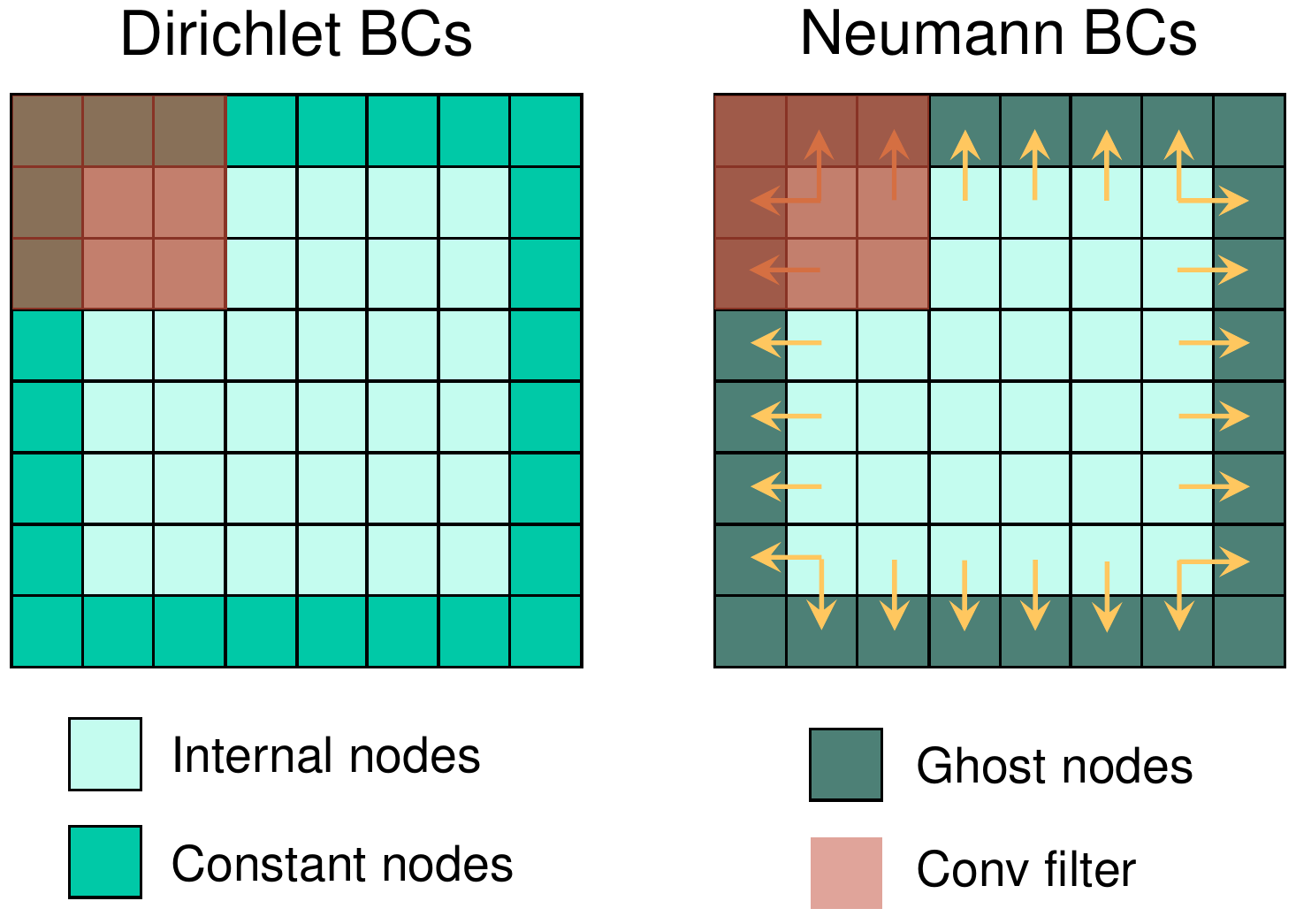} 
\caption{A graphic illustration of hard encoding of BCs into the framework. We perform the hard enforcement of BCs both for NN learning and numerical differentiation.}
\label{fig:BCs}
\end{figure}

Considering the uniform discretization of the spatial domains, it is easy to encode the BCs by using padding pixel-wisely. The graphical depiction is shown in Figure \ref{fig:BCs}. Firstly, for Dirichlet BCs, we rigorously incorporate the known boundary values into the HR tensor $\widehat{\mathbf{u}}^h$. Secondly, for Neumann BCs, we apply ghost nodes \cite{hughes2012finite} for inferring the external node values via finite difference (FD) methods, where the physical correlation between ghost nodes and internal nodes is time-invariant.

\subsection{Physics-informed loss function}
Since SR of scientific data is equation-specific, it is beneficial and necessary to also consider the PDE residual loss apart from the traditional data loss between reconstructed HR variables and the ground-truth HR simulations. Here we apply the numerical differentiation on HR variables $\widehat{\mathbf{u}}^h$ to approximate the derivatives of interest. The discussion on the selection of FD stencils and the introduction of numerical errors is presented in \ref{s:fd_filters}. Furthermore, the specific PDE residual is formulated as $\boldsymbol{\mathcal{R}}(\mathbf{x},t; \boldsymbol{\theta}):=\widehat{\mathbf{u}}_t^h + \mathcal{F}[\widehat{\mathbf{u}}^h; \boldsymbol{\lambda}]$ where $\boldsymbol{\theta}$ is the NN parameters. For a two-dimensional PDE, the physics loss is finally given by
\begin{equation}
    \label{eq:lp}
    \mathcal{L}_p = \frac{1}{N_{x}N_{y}N_{t}N_{b}} \left \|{\boldsymbol{\mathcal{R}}(\mathbf{x},t; \boldsymbol{\theta})}\right\|_2,
\end{equation}
where $N_x,N_y$ are the width and height in the spatial grid; $N_t$ is the number of time steps; $N_b$ is the batch size; $\left \| \cdot \right\|_2$ denotes Frobenius $\ell_2$ norm. Moreover, the data loss is measured by the averaged $\ell_1$ norm between HR ground truth $\mathbf{u}^{*}$ and reconstructed variables $\widehat{\mathbf{u}}^h$, which is formulated as 
\begin{equation}
    \label{eq:ld}
    \mathcal{L}_d = \frac{1}{N_{x}N_{y}N_{t}N_{b}} \left \| \mathbf{u}^{*} -  \widehat{\mathbf{u}}^h \right\|_1.
\end{equation}
Next, the total loss is a weighted combination of data loss $\mathcal{L}_d$ and physics loss $\mathcal{L}_p$ for optimizing the network by both considering the guidance of HR labeled data and the constraint of physical laws. It is given by 
\begin{equation}
    \label{eq:loss}
    \mathcal{L} = \mathcal{L}_d + \beta \mathcal{L}_p,
\end{equation}
where $\beta$ is the weighting coefficient.

\section{Experiments}\label{s:experiment}
In this section, we conduct extensive numerical experiments to evaluate the performance of our proposed approach on different types of spatiotemporal systems, ranging from convection phenomena to reaction-diffusion (RD) dynamics. The specific investigations include: (1) a comparative study on three PDE systems against the baseline models; (2) an ablation study on the effectiveness of ConvLSTM, the introduction of physics loss and the encoding of BCs. The results indicate the superiority of our framework in reconstruction accuracy, computational efficiency and scalability.

\subsection{Setup}\label{s:setup}
\subsubsection{Datasets}\label{s:dataset}
In the numerical experiments, we first consider 2D RBC system \cite{esmaeilzadeh2020meshfreeflownet}, which is a paradigm for nonlinear and chaotic dynamics. The RBC model describes the buoyancy-driven flow of a fluid heated from below and cooled from above and contains a non-trivial coupling between velocity, pressure and temperature. It is one of the most extensively investigated physical phenomena in geophysics, astrophysics, meteorology, oceanography and engineering. The governing equations for RBC system are given by
\begin{equation}\label{eq:rb}
\begin{split}
    &\nabla \cdot \mathbf{u} = \mathbf{0}, \\
    &\frac{\partial \mathbf{u}}{\partial t} + (\mathbf{u} \cdot \nabla) \mathbf{u} = - \nabla p + T \mathbf{e}_x + R^{*} \nabla^2 \mathbf{u}, \\
    &\frac{\partial T}{\partial t} + (\mathbf{u} \cdot \nabla) T = P^{*} \nabla^2 T,
\end{split}
\end{equation}
where $\mathbf{u}=\{u,v\}$ denotes the velocity components in $x$ and $y$ directions and $p,T$ refer to pressure and temperature terms. Here $R^{*}$ and $P^{*}$ are $(P_r/R_a)^{0.5}$ and $(R_aP_r)^{-0.5}$ respectively, where $R_a$ and $P_r$ represent dimensionless Rayleigh and Prandtl numbers with expression of $g\alpha (\Delta T)h^3/(\nu \kappa)$ and $\nu/\kappa$. $\{g, \alpha, \Delta T, h, \nu, \kappa\}$ are gravity acceleration, thermal expansion coefficient, the temperature difference between the top and bottom walls, the length between the plates, the kinematic viscosity and the heat conductivity coefficient. In addition, $\mathbf{e}_x$ denotes the unit vector in $x$ direction. Herein, we generate the HR ground truth dataset by using the open-source code in \cite{esmaeilzadeh2020meshfreeflownet} with $R_a$ and $P_r$ as $1\times10^6$ and $1$ respectively. The spatial domain of $[0,4]\times[0,1]$ is discretized as $512 \times 128$, and the time period is $[0,16]$ for $1600$ steps. Next, we divide the HR dataset into 100 samples with 16 snapshots and a spatial resolution of $512 \times 128$.

Apart from the convection phenomena, we also evaluate the performance of our proposed framework on RD equations. In particular, Gray-Scott (GS) RD equations are considered due to their depictions of complex dynamical patterns and local details, which work as an ideal system to test the predictive limit of the proposed approach \cite{geneva2022transformers}. In general, the GS RD system, which typically portrays complex chemical and biological process, is governed by 
\begin{equation}
    \label{eq:gs}
    \mathbf{u}_t=\mathbf{D} \Delta \mathbf{u} + \mathbf{R}(\mathbf{u})
\end{equation}
where $\mathbf{u}=[u,v]^\top$ denotes the concentration vector, $\mathbf{D}=[\gamma_u,0;0,\gamma_v]$ is the diagonal diffusion coefficient matrix and $\mathbf{R}=[-uv^2+f(1-u),uv^2-(f+k)v]^\top$ represents the reaction vector. Moreover, we conduct the comparative study between our PhySR against baseline models on both 2D and 3D GS RD dynamics. In specific, we set $\{\gamma_u,\gamma_v,f,k\}$ to be $\{0.16,0.08,0.06,0.062\}$ for 2D case and $\{0.2,0.1,0.025,0.055\}$ for 3D case of the GS RD systems. In order to train the network for 2D GS RD dynamics, we generate 400 reference solutions/samples, where the spatial grid size is $256 \times 256$ with $\delta x=1$ and the simulated time period is $[0,200]$ with $\delta t=10$. In light of the intensive computation for 3D GS RD case, we generate 400 reference HR samples with the Cartesian grid of $48\times 48 \times 48$ for $\mathbf{x}\in[0,100]^3$ and 20 time steps for $t\in[0,20]$ (i.e., $\delta t=1$). All of the GS RD datasets are obtained via the high-order FD method with periodic BCs.

Furthermore, the HR labeled datasets are downsampled uniformly in the spatiotemporal domains, and blurred simultaneously for spatial resolutions. Herein, we design two ST-SR scenarios with different downsampling ratio $r_{(\cdot)}$ for evaluation: (1) $r_t=4,r_s=8$ for 2D RBC and 2D GS dynamics; (2) $r_t=2,r_s=4$ for 3D GS system. In addition, we set the splitting ratio for training and testing datasets as 70\% and 30\%.

\subsubsection{Baselines}\label{s:baseline}
Although there are various baseline models for the traditional video ST-SR task, they are not specifically designed for ST-SR of scientific data, which may cause the unfairness in the experimental evaluation. To this end, here we consider the state-of-the-art baseline model MeshfreeFlowNet \cite{esmaeilzadeh2020meshfreeflownet} in the field of scientific ST-SR for the performance comparison. In addition, we consider the classical interpolation approach (i.e., trilinear interpolation for 2D PDEs and quadlinear interpolation for 3D PDE case) as a benchmark method for comparison.

Regarding the MeshfreeFlowNet, we adopt the same network architecture\footnote{Except the number of the input and output channels are set to be 2 due to the state variables of $u$ and $v$ in our 2D and 3D GS RD datasets.} as in \cite{esmaeilzadeh2020meshfreeflownet}, i.e., a convolutional encoder named the Context Generation Network to map the LR inputs to a latent context grid, followed by a fully connected neural network (FCNN) called Continuous Decoding Network to generate continuous spatiotemporal solutions. The Context Generation Network is implemented with 3D variant of the U-Net \cite{ronneberger2015u} with residual blocks. The loss function used for training the MeshfreeFlowNet also follows the form of Eq. \eref{eq:loss}, i.e., a linear combination of data (or regression) and physics loss. The weighting coefficient $\beta$ for 2D RBC system is set as 0.05 which is the optimal value reported in \cite{esmaeilzadeh2020meshfreeflownet}, and we select $\beta=0.025$ for 2D GS case after empirical tests with $\{0.01,0.025,0.05,0.1,0.5,1,5,10\}$. It should be noted that the physics loss in MeshfreeFlowNet is computed via automatic differentiation since the fully connected Continuous Decoding Network is used for generating the HR solution. The training is conducted on four NVIDIA Tesla V100 GPU cards (32G) due to the limit of our computational resources.

\subsubsection{Evaluation metrics}\label{s:metrics}
In this paper, we consider using relative full-field $\ell_2$ error $\epsilon$ to measure the discrepancy between the ground truth HR dynamics $\mathbf{u}^{*}$ and the reconstructed HR variables $\widehat{\mathbf{u}}^{h}$. Specifically, it has the formulation of  
\begin{equation}
    \label{eq:error}
    \epsilon = \sqrt{\frac{\left \|\mathbf{u}^{*}-\widehat{\mathbf{u}}^{h}\right\|_2}{\left \|\mathbf{u}^{*}\right\|_2}} \times 100\%.
\end{equation}

Moreover, we also compare our PhySR with two baseline models on other three metrics, including (1) the number of trainable parameters, (2) the training time $T_\text{train}$ and (3) the inference time $T_\text{infer}$. These metrics are employed for evaluating computational and memory efficiency.

\subsubsection{Implementation}\label{s:implement}
Specifically, in this paper, we consider periodic BCs for the numerical experiments of 2D and 3D GS RD equations where periodic padding (i.e., circular padding in Pytorch \cite{paszke2019pytorch}) is adopted in our network both for feature mapping and HR outputs. For the RBC system, the boundary information is unknown but we also apply periodic padding for feature mapping. The similar padding operation appears in another forecasting work of RBC system \cite{wang2020towards}.

We train our PhySR using the Adam optimizer \cite{kingma2014adam} with 1000 epochs for 2D RBC system and 4000 epochs for both 2D and 3D GS RD cases. The learning rate is set as 0.001 and the weight decay is $1\times 10^{-6}$. The batch size is defined as 16. We use 2 residual blocks with the feature channel of 32 for both 2D and 3D cases. The size of the convolution kernel is set to be 3. Note that the weighting coefficient $\beta$ in Eq. \eref{eq:loss} is set as 0.025 for all three PDE systems after empirical trials of $\{0.01,0.025,0.05,0.1,0.5,1,5,10\}$. While training the PhySR models, we normalize the input LR measurements by the mean and standard deviation of the HR ground truth. All of the PhySR models are implemented with PyTorch and trained on a NVIDIA Tesla V100 GPU card (32G) in a standard workstation.

\subsection{Comparative study}\label{s:compare}
We compare the ST-SR performance between PhySR and the baselines by using three evaluation metrics as shown in Section \ref{s:metrics} and the reconstructed HR snapshots. The main finding is that our PhySR has much fewer parameters compared with MeshfreeFlowNet but holds the best reconstruction accuracy among all applicable methods. Moreover, PhySR is computationally efficient, which is consistently faster than MeshfreeFlowNet in both training and inference. Regarding the classical interpolation methods (i.e., trilinear and quadlinear), the expansion of dimensions induces an significant increase in inference time, which makes it undesirable for recovering high-dimensional complex systems.

\begin{table}[t!]
\centering
\caption{The comparison between PhySR and baseline models. ``\# Parameters'' refers to the number of trainable parameters. $T_{\text{train}}$ and $T_{\text{infer}}$ denote the training time per epoch and the inference time per time step, respectively. For MeshfreeFlowNet and PhySR, means and standard deviations are presented by performing 10 training times with different random seeds. Besides, ``N/A'' represents being inapplicable to the specific index, and ``-'' means failure for being out of memory due to 4D convolution. Note that ``M'' denotes million and ``ms'' is millisecond.}
{\small
\begin{tabular}{c c c c c c}
\toprule
PDEs & Methods & {\# Parameters [M]} & {$T_{\text{train}}$ [s]} & {$T_{\text{infer}}$ [ms]} & Relative full-field error [\%] \\
\hline
\multirow{3}{*}{2D RBC} & Interpolation & N/A & N/A & $\mathbf{0.095}$ & 20.78 \\ \cline{2-6}
& MeshfreeFlowNet & 6.95 & 15.86 & 56.10 & $4.39\pm0.32$ \\ \cline{2-6}
& PhySR & $\mathbf{0.29}$ & $\mathbf{4.70}$ & 11.15 & $\mathbf{2.17\pm0.26}$ \\ 
\hline
\multirow{3}{*}{2D GS} & Interpolation & N/A & N/A & $\mathbf{0.08}$ & 31.58 \\ \cline{2-6}
& MeshfreeFlowNet & 4.28 & 27.71 & 56.06 & $18.31 \pm 1.87$ \\ \cline{2-6}
& PhySR & $\mathbf{0.23}$ & $\mathbf{6.91}$ & 9.64 & $\mathbf{3.60\pm0.27}$ \\ 
\hline
\multirow{3}{*}{3D GS} & Interpolation & N/A & N/A & 364.91 & 31.42 \\ \cline{2-6}
& MeshfreeFlowNet & -- & -- & -- & -- \\ \cline{2-6}
& PhySR & $\mathbf{0.68}$ & $\mathbf{12.13}$ & $\mathbf{12.68}$ & $\mathbf{2.77\pm0.19}$ \\ 
\bottomrule
\end{tabular}
}
\label{tab:comp_train}
\end{table}

\subsubsection{2D Rayleigh-B{\'e}nard convection system}\label{s:2drbc}
The first case is the 2D RBC system. From Table \ref{tab:comp_train}, we can see that both PhySR and MeshfreeFlowNet predict the HR dynamics well and achieve excellent reconstruction accuracy ($\epsilon$ under $5\%$), while the traditional interpolation method (i.e. trilinear interpolation) presents a non-negligible error ($\epsilon > 20\%$) and exhibits obvious weakness in high-dimensional upsampling. In addition, the slight superiority of PhySR is rooted in the explicit modeling of temporal evolution by introducing the ConvLSTM architecture to PhySR. Nevertheless, MeshfreeFlowNet directly applies 3D convolution operations to realize the temporal upscaling.

\begin{figure}[t]
\centering
\includegraphics[width=0.99\columnwidth]{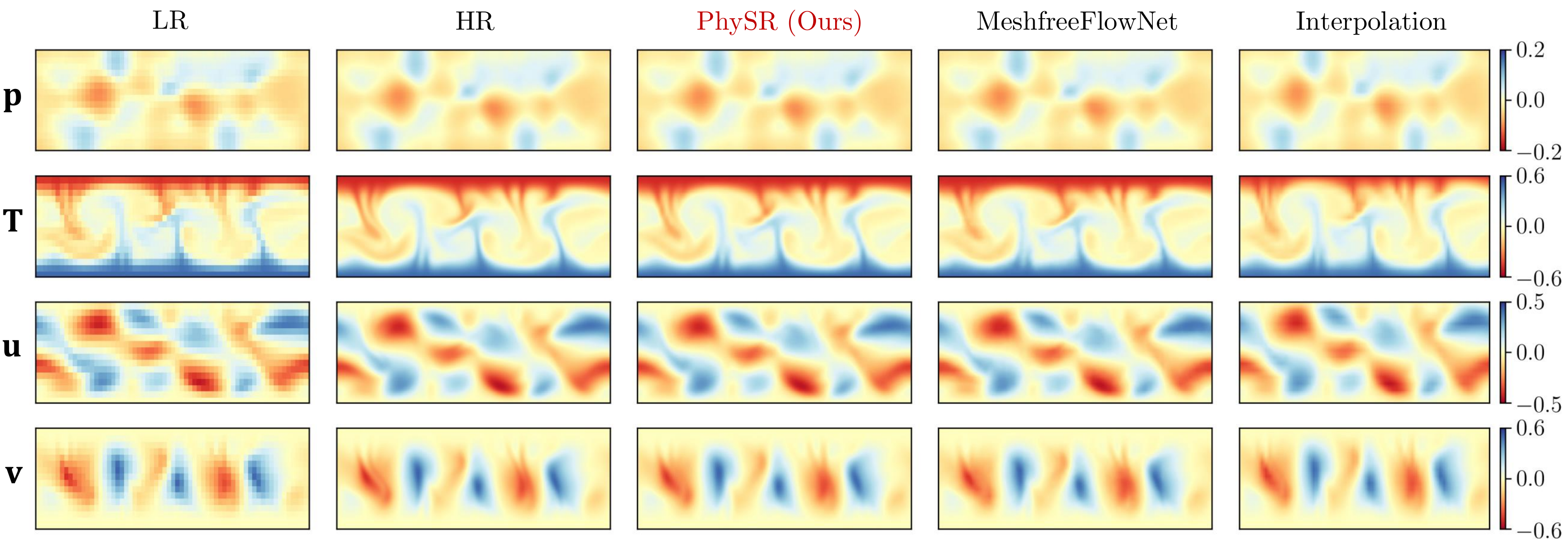} 
\caption{The comparison of SR performance for 2D RBC system at representative snapshots, including four physical quantities (i.e., $p,T,u,v$). The corresponding Rayleigh and Prandtl numbers are $1\times 10^6$ and $1$. The spatial domain is $[0,4]\times[0,1]$, and the specific snapshot is selected at time $t=14.4$.}
\label{fig:comp_2drb}
\end{figure}

Moreover, we compare the snapshots of these ST-SR results to further understand the reconstructed spatiotemporal dynamics. As shown in Figure \ref{fig:comp_2drb}, we observe the consistent results that PhySR presents excellent ST-SR performance and the reconstruction of MeshfreeFlowNet is also very close to the HR ground truth. The interpolation method can generally capture the dynamical behaviors but produces more vague reconstruction compared with the other two methods. It is also noteworthy that MeshfreeFlowNet has a larger number of network parameters, which causes longer training and inference time compared with PhySR. Overall, the comparative study on 2D RBC case validates the effectiveness and efficiency of PhySR.

\subsubsection{2D GS equation}\label{s:2dgs}
RD dynamics are also tested for measuring the ST-SR performances of PhySR and baseline models in addition to the convection system. The second PDE system, 2D GS RD equation, also holds complex and rich dynamical patterns. The characteristic of 2D GS system lies in the delicate local features, which poses challenges to the network training. As presented in Table \ref{tab:comp_train}, it is evident that the relative full-field errors of MeshfreeFlowNet and trilinear interpolation models increase largely. It empirically indicates their limitations to such complicated spatiotemporal systems with various local details. Nevertheless, PhySR still holds the satisfactory ST-SR performance for 2D GS RD system though the error $\epsilon$ rises up to $3.60\pm0.27\%$ due to the complexity of the dynamical behaviors.

\begin{figure}[t]
\centering
\includegraphics[width=0.9\columnwidth]{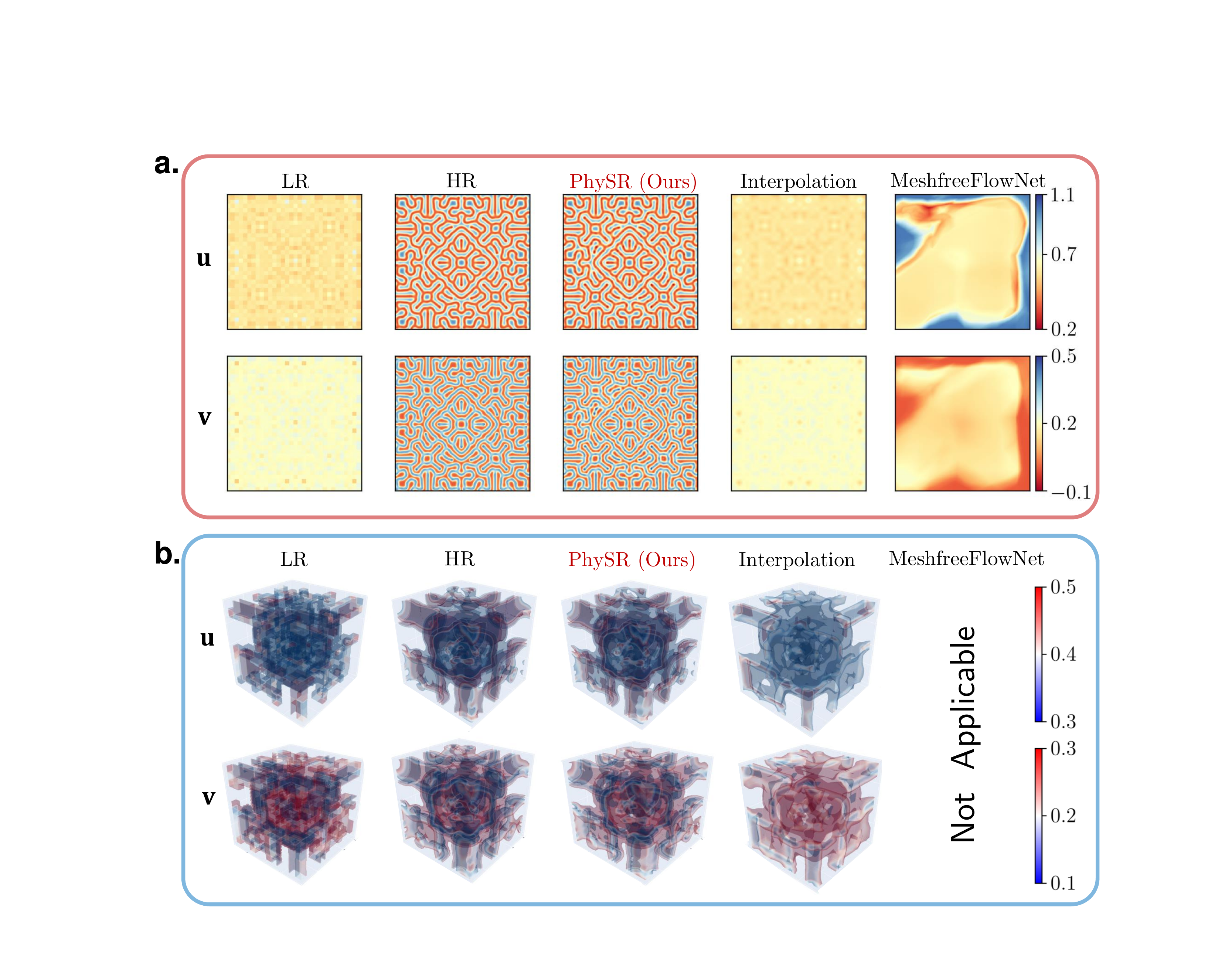} 
\caption{The comparison of representative snapshots for GS RD systems. Here (a) and (b) represent the results for 2D and 3D GS equations respectively. The spatial domains are $[0,256]^2$ for (a) and $[-50,50]^2$ for (b). All of the snapshots are selected at time $t=0.5$. Note that MeshfreeFlowNet for (b) is unavailable due to out-of-memory.}
\label{fig:comp_gs}
\end{figure}

From Figure \ref{fig:comp_gs}(a), we find that the discrepancy between LR measurement data and HR ground truth is significant, which also implies the difficulty of reconstructing GS RD systems. Correspondingly, PhySR is still capable of portraying general patterns and evolutionary details of 2D GS dynamics, whereas MeshfreeFlowNet and trilinear barely discover the clear patterns. We ascribe the unsatisfactory results of MeshfreeFlowNet to two-folds: (1) error propagation due to lack of explicit temporal modeling; (2) the global learning scheme of FCNN module which tends to fail in capturing delicate local dynamics \cite{gao2021phygeonet,ren2022phycrnet}. The experiment on 2D GS RD equation further validates the superiority of PhySR over baseline models thanks to the specified time marching strategy.

\subsubsection{3D GS equation}\label{s:3dgs}
Thirdly, we consider a higher dimensional RD system (i.e., 3D GS equation) to evaluate the ST-SR performance of PhySR against two baseline methods. Note that MeshfreeFlowNet requires demanding computational memory for the 3D GS case by using 4D convolutional operations, which exceeds the limit of our server. Therefore, we mark it a failed test for MeshfreeFlowNet. In addition, we observe that the high-dimensional linear interpolation method (i.e., quadlinear) shows much slower inference performance compared with those for 2D cases, and the reconstruction results are still unsatisfactory. However, PhySR showcases outstanding performance regarding all these three challenging requirements: reconstruction accuracy, computational efficiency and scalability. As shown in Figure \ref{fig:comp_gs}(b), the representative predicted snapshots of PhySR match the corresponding HR ground truth well, which proves its remarkable capability for ST-SR of scientific data even in higher dimensions. We can also see the positive evidences of PhySR in Table \ref{tab:comp_train} that the reconstruction accuracy is appealing ($\epsilon$ under $5\%$) and the training and inference time increase mildly when extended to the 3D case.

\subsection{Ablation Study}\label{s:ablation}
To evaluate the effectiveness of each component of our PhySR framework, we perform the ablation study on 2D GS equation under ST-SR scenarios $S_1$ (shown in \ref{s:dataset}). Here we mainly investigate the contributions of ConvLSTM for temporal propagation of dynamics, the introduction of physics loss $\mathcal{L}_p$ and the hard-encoding of BCs. These three components are specialized for ST-SR of scientific data, which makes them worthwhile to be further analyzed. Note that the hyper-parameters (i.e., $\beta=0.025$) and the training effort are kept same for all numerical experiments in this part.

\subsubsection{Effectiveness of ConvLSTM}\label{s:ablation_convlstm}
Firstly, we examine the importance of ConvLSTM for temporal refinement. We consider two models: (A) the full PhySR with ConvLSTM, and (B) PhySR without ConvLSTM where the corresponding part is substituted by a standard convolution operation with the same channel size. From Figure \ref{fig:ablation}, we observe that the reconstruction snapshot of Model (B) clearly presents much blurrier results than that of Model (A). It indicates the relatively weaker capability of Model (B) in capturing the temporal evolution compared with Model (A) due to the lack of the explicit time marching module (i.e., ConvLSTM). Besides, the comparison of relative full-field errors between Model (A) and (B) in Table \ref{tab:ablation} further validates that the incorporation of ConvLSTM effectively contributes to temporal refinement and enhances the performance of temporal upsampling.

\subsubsection{Effectiveness of physics loss $\mathcal{L}_p$}\label{s:ablation_lp}
Secondly, to test the effectiveness of introducing physics loss $\mathcal{L}_p$, we design two comparative experiments for ST-SR of scientific data. We consider two models: (A) the full PhySR constrained by both data loss $\mathcal{L}_d$ and physics loss $\mathcal{L}_p$, and (C) PhySR only trained with data loss $\mathcal{L}_d$. As shown in Figure \ref{fig:ablation}, the reconstruction of Model (C) presents relatively discontinuous contours (marked in black boxes) compared with the prediction snapshot of Model (A). This vague phenomenon is induced by the lack of the constraint of physical laws. Namely, embedding physical principles into the learning framework can facilitate the interpolation/extrapolation ability in ST-SR. In addition, we also show that the full PhySR model outperforms Model (C) in reconstruction accuracy (see Table \ref{tab:ablation}), which further validates the significant of physics loss for ST-SR of scientific data.

\subsubsection{Effectiveness of hard-encoding of BCs}\label{s:ablation_bc}
\begin{figure}[t]
\centering
\includegraphics[width=0.9\columnwidth]{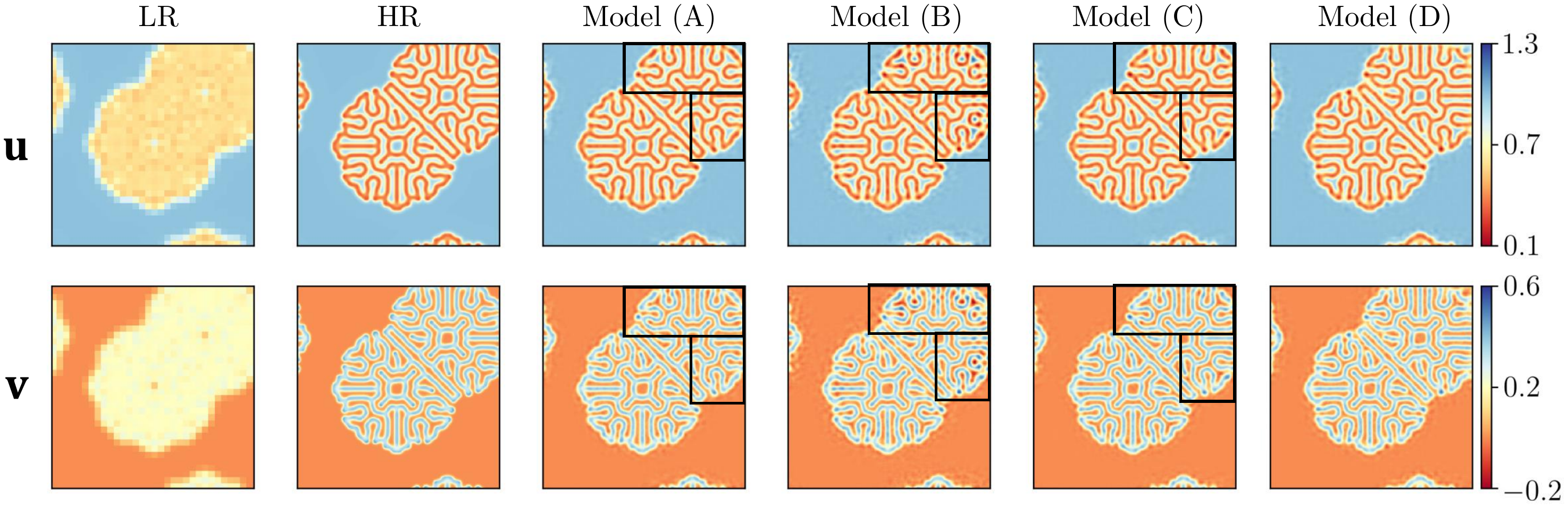} 
\caption{The comparison of representative snapshots on ablation study. The evaluation is performed on 2D GS RD system. The spatial domain is $[0,256]^2$.}
\label{fig:ablation}
\end{figure}

Thirdly, we evaluate the effectiveness of hard-encoding BCs in the network. There are also two models designed for comparison: (A) the full PhySR with BCs encoded, and (D) PhySR without BCs incorporated. In this part, Model (D) is trained by replacing periodic padding to zero padding in feature mapping and does not consider the known BCs enforced into the output variables. It is apparent that the reconstructed results on the boundaries of Model (D) are not competitive as those of Model (A), especially on the top side (see Figure \ref{fig:ablation}). Moreover, as shown in Table \ref{tab:ablation}, we observe the consistent performances that the ST-SR accuracy of Model (D) is inferior than that of Model (A). The effectiveness of encoding BCs is also experimentally validated. Hence, it is beneficial to incorporate BCs into learning machines when the boundary information is available.

\begin{table}[t!]
\centering
\caption{The comparative results of ablation study on 2D GS RD equations. Means and standard deviations are computed via 10 training times with different random seeds.}
{\small
\begin{tabular}{c c c c c}
\toprule
Metric & Model (A) & Model (B) & Model (C) & Model (D) \\
\hline
Relative full-field error [\%] & $3.60\pm0.27$ & $4.49\pm0.36$ & $3.71\pm0.32$ & $4.32\pm0.42$ \\
\bottomrule
\end{tabular}
}
\label{tab:ablation}
\end{table}

\section{Discussion}\label{s:discussion}
In this section, we discuss the advantages and current limitations of the proposed discrete learning framework for ST-SR, as well as the possible solutions for future research directions. Firstly, there are three main benefits of our physics-informed discrete learning architecture, i.e., excellent reconstruction accuracy, efficiency and scalability. Regarding the ST-SR accuracy, PhySR outperforms baseline models due to the specialized time marching neural module (e.g., ConvLSTM) for learning temporal evolution and the incorporation of physical principles (i.e., governing equations and boundary information). Although there is a concern about the numerical errors caused by FD approximation of derivatives compared with the automatic differentiation through fully-connected NNs \cite{raissi2019physics,esmaeilzadeh2020meshfreeflownet}, we show that high-order FD kernels can be utilized to mitigate this issue in discrete learning frameworks, as reported in \ref{s:fd_filters}. The reconstruction accuracy is improved when a higher-order FD kernel is chosen. In addition, we decompose the ST-SR task into temporal upsampling and spatial reconstruction. This setting helps to make the model scalable to higher-dimensional PDE systems (e.g., 3D GS RD equation) compared with the network directly using 3D/4D convolution. Moreover, in the part of spatial reconstruction, the spatial features are learned in LR latent space. Thus, a small convolutional filter (e.g., filter size of 3) can be applied for feature mapping, which facilitates the efficiency and reduces the computational memory. Although the performance of PhySR is tested on convection and RD systems, it is also applicable to other spatiotemporal systems where the PDE formulation is given. 

However, there still remain several issues that are worthy to explore in the future. For example, the present network is rooted in convolution operation and limited to structured grids of computational domains. Leveraging graph neural networks (GNNs) is an alternative to resolve this problem, but it is challenging to determine the derivative terms based on irregular meshes and further assemble the PDE residuals. Secondly, the requirement for time stepping is demanding when dealing with very chaotic or fast-evolving systems because of the nature of discretized methods. We need to define a relatively smaller time interval ($\delta t$) for training compared with continuous learning architectures. Thirdly, the physics-informed learning essentially considers the physical laws as a soft regularizor, which makes the optimization and hyper-parameters tuning intractable. To handle this, encoding the PDE structures into the network can be a promising direction for learning a model that strictly obeys the underlying governing equations \cite{rao2022discovering,rao2021embedding}.

\section{Conclusion}\label{s:conclusion}
In this paper, we propose a novel and efficient physics-informed deep super-resolution framework (PhySR) for spatiotemporal scientific data. The network is comprised of the temporal interpolation, temporal refinement and spatial reconstruction modules, which is inspired by the independence between spatial and temporal derivative terms in a general PDE form. Specifically, we utilize the interpolation to directly increase the frame rate and ConvLSTM layers to correct the errors of temporal upampling. Furthermore, we adopt residual blocks to extract LR features, sub-pixel layers for spatial upscaling and a global residual connection for stable propagation of spatial information. Besides, the BCs are hard-encoded into our network to facilitate convergence and improve the approximation on boundaries. The performance of our proposed framework has been assessed on three nonlinear PDE systems (i.e., 2D RBC system, 2D GS and 3D GS RD equations) and compared with baseline algorithms. The numerical results show the superiority of our method in reconstruction accuracy, efficiency and scalability. Moreover, the ablation study validates the effectiveness of the incorporation of ConvLSTM and physics principles.

\section*{Acknowledgement}
P. Ren would like to gratefully acknowledge the Vilas Mujumdar Fellowship at Northeastern University. Y. Liu would like to thank the support from the Fundamental Research Funds for the Central Universities. The work is also supported by the Beijing Outstanding Young Scientist Program (No. BJJWZYJH012019100020098) as well as the Intelligent Social Governance Platform, Major Innovation \& Planning Interdisciplinary Platform for the ``Double-First Class'' Initiative, Renmin University of China.

\section*{Data Availability}
All the datasets and source codes to reproduce the results in this study are available on GitHub at \url{https://github.com/paulpuren/PhySR}.

\appendix
\section{Selection of FD kernels}\label{s:fd_filters}

FD kernels are used to approximate derivative terms in PDEs, which directly affect the computational efficiency and the reconstruction accuracy. Therefore, it is crucial to choose appropriate FD kernels for discretized-based learning frameworks. For spatiotemporal systems, we need to both consider temporal and spatial derivatives. In specific, second-order central difference is utilized for calculating temporal derivatives, i.e.,
\begin{equation}
    \label{eq:approx_dt}
    \frac{\partial u}{\partial t} = \frac{-u(t - \delta t, \xi, \eta) + u(t +\delta t, \xi, \eta)}{2 \delta t} + \mathcal{O}((\delta t)^2), 
\end{equation}
where $\{\xi,\eta\}$ represent the spatial locations and $\delta t$ is time spacing. In the network implementation, it can be organized as a convolutional kernel $K_t$, 
\begin{equation}
    \label{eq:kernel_dt}
    K_t = [-1,0,1] \times \frac{1}{2\delta t}.
\end{equation}
However, the derivatives at first and last steps are inaccessible due to the nature of central difference scheme. Herein, we substitute it with forward and backward Euler methods to compute the corresponding time derivatives (at first and last steps).

Likewise, we also apply central difference to calculate the spatial derivatives for internal nodes and use forward/backward difference for boundary nodes. For instance, in this paper, the fourth-order central difference is utilized to approximate the first and second spatial derivatives. The FD kernels for 2D cases with the shape of $5\times 5$ are given by 
\begin{equation}
    \label{eq:kernel_dx}
    K_{s,1} = {
    \begin{bmatrix} 
        0 & 0 & 0 & 0 & 0 \\
        0 & 0 & 0 & 0 & 0 \\
        1 & -8 & 0 & 8 & -1 \\
        0 & 0 & 0 & 0 & 0 \\  
        0 & 0 & 0 & 0 & 0 
    \end{bmatrix}} \times \frac{1}{12(\delta x)}, \;\;   
    K_{s,2} = {
    \begin{bmatrix} 
        0 & 0 & -1 & 0 & 0 \\
        0 & 0 & 16 & 0 & 0 \\
        -1 & 16 & -60 & 16 & -1 \\
        0 & 0 & 16 & 0 & 0 \\  
        0 & 0 & -1 & 0 & 0 
    \end{bmatrix}} \times \frac{1}{12(\delta x)^2},
\end{equation}
where $\delta x$ denotes the grid size of HR variables; $K_{s,1}$ and $K_{s,2}$ are FD kernels for the first and second derivatives respectively. In addition, we conduct a parametric study on the selection of FD kernels, including the second-order ($3\times 3$), the fourth-order ($5\times 5$) and the sixth-order ($7\times 7$) central difference strategies. The numerical experiments are based on 2D GS RD system. As shown in Table \ref{tab:comp_kernels}, the training time rises up and the reconstruction accuracy grows as the FD order increases. Nevertheless, we also notice that the PhySR with $5\times5$ FD kernel can already produce competitive results compared with the model using $7\times7$ FD kernel. It reflects that the reconstruction performance can be improved very slowly with a higher-order FD scheme (e.g., sixth-order) after the approximation accuracy of derivatives are satisfied. Therefore, we select the $5\times5$ FD kernel in this paper for the numerical experiments considering a trade-off between accuracy and efficiency. Generally, for those systems with complex patterns or fast-evolving features (e.g., the chaotic RBC system), it is more suitable to consider high-order FD kernels.

\begin{table}[t!]
\centering
\caption{The comparative results on the selection of FD kernels. $T_{\text{train}}$ denotes the training time per epoch. The experiments are conducted on 2D GS RD equations. For each scenario, we train the network 10 times with different random seeds to obtain the corresponding mean and standard deviation.}
{\small
\begin{tabular}{c c c}
\toprule
FD kernels & {$T_{\text{train}}$ [s]} & Relative full-field error [\%] \\
\hline
$3\times3$ & 6.85 & $3.68\pm0.27$ \\
\hline
$5\times5$ & 6.91 & $3.60\pm0.27$ \\
\hline
$7\times7$ & 7.16 & $3.60\pm0.23$ \\
\bottomrule
\end{tabular}
}
\label{tab:comp_kernels}
\end{table}



\bibliographystyle{elsarticle-num}
\bibliography{refs.bib}

\end{document}